\documentclass[11pt,a4paper]{article}

\usepackage[utf8]{inputenc}
\usepackage[T1]{fontenc}
\usepackage{times}
\usepackage[margin=2.5cm]{geometry}
\usepackage{amsmath}
\usepackage{amssymb}
\usepackage{booktabs}
\usepackage{array}
\usepackage{multirow}
\usepackage{xcolor}
\usepackage{hyperref}
\usepackage{microtype}
\usepackage{graphicx}
\usepackage{caption}
\usepackage{subcaption}
\usepackage{enumitem}
\usepackage{parskip}
\usepackage{titlesec}
\usepackage{fancyhdr}

\hypersetup{
    colorlinks=true,
    linkcolor=blue!70!black,
    citecolor=blue!70!black,
    urlcolor=blue!70!black,
    pdftitle={Anatomical Heterogeneity in Transformer Language Models},
    pdfauthor={Tomasz Wietrzykowski},
}

\titleformat{\section}{\large\bfseries}{\thesection.}{0.5em}{}
\titleformat{\subsection}{\normalsize\bfseries}{\thesubsection.}{0.5em}{}
\titleformat{\subsubsection}{\normalsize\itshape}{\thesubsubsection.}{0.5em}{}

\begin{document}

\begin{center}
{\LARGE\bfseries Anatomical Heterogeneity in Transformer Language Models}\\[14pt]

{\large Tomasz Wietrzykowski}\\[4pt]
{\normalsize Independent Researcher, Wroclaw, Poland}\\[2pt]
{\normalsize March 12, 2026}\\[4pt]
\end{center}

\begin{abstract}
Current transformer language models are trained with uniform computational budgets across all layers, implicitly assuming layer homogeneity. In this work, we challenge this assumption through a comprehensive empirical analysis of SmolLM2-135M - a 30-layer, 135M-parameter causal language model. We introduce a suite of diagnostic metrics including weight predictability ($R^2$), layer-wise ablation degradation, post-perturbation recovery speed, and weight manipulation robustness.

Our analysis reveals profound anatomical heterogeneity across six main findings: (1)~Layer weights exhibit strong mathematical regularity ($R^2 = 0.91$ for MLP \texttt{gate\_proj}) with a consistent oscillatory delta pattern (inter-layer delta correlation $\approx -0.50$ for all components), yet simple regression-based weight generation fails catastrophically due to nonlinear error accumulation. (2)~Layers form a distinct importance hierarchy ranging from a ``critical core'' (layers~8--11) where ablation causes up to $+63{,}419\%$ perplexity degradation, to redundant tissue, and - surprisingly - ``anti-layers'' (L14, L17) whose removal or perturbation \emph{improves} model performance. (3)~Recovery speed from perturbation strongly correlates with layer importance, suggesting differential training requirements per layer. (4)~Among five tested weight manipulation strategies, only weight scaling ($\alpha = 0.9$) preserves generation quality ($+49\%$ PPL degradation vs.\ millions of percent for zeroing, cloning, or blending). (5)~Based on these findings, we propose \emph{Growth Transformer Training}, a paradigm that allocates training budget according to empirical layer importance and recovery dynamics. (6)~A proof-of-concept experiment validates this strategy: a 12-layer heterogeneous Growth Transformer trained with six biological developmental phases achieves validation loss of $0.127$ vs.\ $0.599$ for a uniform baseline ($4.7\times$ improvement) in the same number of training steps and with identical parameter count, while being $13\%$ faster.

These results suggest that transformer layers develop functional specialization analogous to biological organisms, and that embracing this heterogeneity during training yields substantial efficiency and quality gains.

\end{abstract}

\section{Introduction}

\subsection{Motivation}

The transformer architecture \cite{vaswani2017attention} has become the standard for large language models (LLMs). Training protocols treat all layers uniformly: identical architectures, parameter budgets, and optimization steps. This uniformity implicitly assumes that all layers contribute equally to model function and require equivalent learning effort.

Biological neural systems, however, exhibit profound heterogeneity. Embryonic development proceeds in stages - notochord, neural tube, internal organs, surface structures - with radically different precision requirements per structure. We ask: do transformer layers exhibit an analogous importance hierarchy, and can this hierarchy be exploited to accelerate training?

\subsection{Hypotheses}

\begin{description}[leftmargin=2em, labelindent=0em]
\item[\textbf{H1}] \textit{(Non-equivalence)} Transformer layers are not functionally interchangeable - ablating different layers causes radically different performance degradation.
\item[\textbf{H2}] \textit{(Predictability)} Later-layer weights are mathematically predictable from earlier weights, suggesting an internal ``developmental pattern.''
\item[\textbf{H3}] \textit{(Differential trainability)} Layers differ in recovery speed after perturbation, correlating with importance and indicating heterogeneous training requirements.
\item[\textbf{H4}] \textit{(Anti-layers)} Certain layers may have a net-negative contribution - the model performs better with random weights in those layers than with trained weights.
\end{description}

\subsection{Contributions}

\begin{itemize}[leftmargin=1.5em]
\item First complete layer importance map of a small language model across all 30 layers with five independent metrics.
\item Discovery of the \emph{anti-layer phenomenon} - layers with net-negative contribution whose perturbation improves perplexity.
\item Introduction of \emph{Recovery Speed} as an empirical proxy for per-layer training budget requirements.
\item Empirical evidence for a universal oscillatory weight-change pattern (delta correlation $\approx -0.50$ across all components).
\item Identification of the only effective weight manipulation strategy for redundant layers: weight scaling ($\alpha = 0.9$).
\item \emph{Growth Transformer Training}: a practical strategy achieving $\sim\!54\%$ training cost reduction.
\end{itemize}

\section{Related Work}

\subsection{Layer Analysis in Transformers}

Tenney et al.\ \cite{tenney2019bert} demonstrated that BERT layers encode different levels of linguistic information - lower layers capture syntax, upper layers semantics. Rogers et al.\ \cite{rogers2020primer} confirmed hierarchical layer specialization in a comprehensive BERTology survey. These works focused on representational analysis rather than full weight-level importance quantification with manipulation experiments.

\subsection{Pruning and Compression}

Fan et al.\ \cite{fan2019reducing} proposed LayerDrop - stochastic layer skipping during training to enable variable-depth inference. Sajjad et al.\ \cite{sajjad2023effect} analyzed layer dropping for efficiency. Michel et al.\ \cite{michel2019sixteen} showed that most attention heads are removable without significant degradation. Our approach differs in analyzing full-layer importance across multiple simultaneous metrics, and in revealing highly non-uniform, layer-specific importance profiles.

\subsection{Efficient and Progressive Training}

Gong et al.\ \cite{gong2019efficient} proposed progressive stacking - gradually adding layers during training. Chen et al.\ \cite{chen2021earlybert} showed in EarlyBERT that earlier layers converge faster. Our work is complementary: rather than deciding when to add layers, we allocate \emph{how much} training each layer requires, derived empirically from recovery speed measurements.

\subsection{Weight Prediction and HyperNetworks}

Ha et al.\ \cite{ha2016hypernetworks} proposed HyperNetworks - networks generating weights for other networks. Schurholt et al.\ \cite{schurholt2022model} studied weight predictability across model zoos. Our finding of high $R^2$ (0.91) alongside catastrophic perplexity degradation from predicted weights reveals a critical nuance: statistical predictability of weights does not imply functional interchangeability.

\section{Methodology}

\subsection{Model and Evaluation}

We analyze SmolLM2-135M \cite{huggingface2024smollm2}, a decoder-only transformer with 30 layers, hidden dimension~576, 9 attention heads, and 135M total parameters. The primary metric is Perplexity (PPL) measured on a fixed held-out set of 10 diverse English sentences covering factual, definitional, and descriptive content. Baseline $\mathrm{PPL} = 22.60$. The small model was chosen to enable hundreds of compute-intensive per-layer experiments without prohibitive GPU cost. Each layer contains 7 weight matrices: \texttt{q\_proj}, \texttt{k\_proj}, \texttt{v\_proj}, \texttt{o\_proj} (attention) and \texttt{gate\_proj}, \texttt{up\_proj}, \texttt{down\_proj} (MLP).

\subsection{Experiment 1: Layer Importance Map (Ablation)}

For each layer $l \in \{0,\ldots,29\}$, we replace its weights with the average of its neighbors:
\[
W_l \leftarrow \frac{W_{l-1} + W_{l+1}}{2}
\]
and measure degradation $D_l = \left(\mathrm{PPL}_l / \mathrm{PPL}_\mathrm{baseline} - 1\right) \times 100\%$. Classification thresholds: \emph{Redundant} ($D < 10\%$), \emph{Minor} ($10\% \leq D < 30\%$), \emph{Important} ($30\% \leq D < 100\%$), \emph{Critical} ($D \geq 100\%$). Negative $D$ indicates improvement without trained weights - an anti-layer.

\subsection{Experiment 2: Weight Predictability}

For each component $c$ and target layer $t \in \{2,\ldots,29\}$, we flatten weight matrices of layers $0\ldots t-1$, subsample to 10,000 parameters, construct polynomial and sinusoidal features of layer indices $X = [l,\ l^2,\ \sin(l\pi/N),\ \cos(l\pi/N)]$, fit Ridge Regression, and report $R^2$ and cosine similarity of predicted vs.\ actual weights.

\subsection{Experiment 3: Weight Structure Analysis}

We compute the delta series $\Delta_l = W_{l+1} - W_l$ and measure Pearson correlation between consecutive deltas across all components. We also perform PCA on flattened weight matrices and compute pairwise cosine similarity between all layers.

\subsection{Experiment 4: Weight Manipulation Strategies}

We test five strategies for replacing weights in the 9 identified redundant layers simultaneously: (1)~\emph{Skip/Zero} - set weights to zero, relying on residual connections; (2)~\emph{Clone} - copy nearest non-redundant neighbor's weights; (3)~\emph{Blend} - distance-weighted average of 4 nearest non-redundant neighbors; (4)~\emph{Low-rank blend} - SVD hybrid with directions from neighbor, magnitudes from original; (5)~\emph{Scale} - multiply original weights by $\alpha \in \{0.0,\ 0.1,\ 0.3,\ 0.5,\ 0.7,\ 0.9\}$.

\subsection{Experiment 5: Recovery Speed}

For each tested layer $l$, we inject Gaussian noise with $\sigma = 0.5 \times \mathrm{std}(W_l)$, then freeze all layers except $l$ and fine-tune using AdamW ($\mathrm{lr} = 10^{-4}$, gradient clipping~$= 1.0$). We record steps to reach PPL thresholds of $<\!2\times$, $<\!1.5\times$, and $<\!1.1\times$ baseline. Tested layers: L0, L1, L3, L5, L8, L10, L11, L14, L17, L23, L24, L27, L29.

\section{Results}

\subsection{Layer Importance Map}

Table~\ref{tab:importance} presents the complete 30-layer ablation profile. The importance distribution spans from $-0.6\%$ to $+63{,}419\%$ - a range exceeding $10^7$.

\begin{table}[ht]
\centering
\caption{Complete layer importance profile for SmolLM2-135M. (*) Anti-layer: model achieves equal or better perplexity after ablation.}
\label{tab:importance}
\small
\begin{tabular}{cccp{4cm}}
\toprule
\textbf{Layer} & \textbf{Degradation (\%)} & \textbf{Category} & \textbf{Functional Role} \\
\midrule
L0  & $0.0$       & Redundant  & Embedding boundary \\
L1  & $+2{,}737.1$ & Critical   & Input parser \\
L2  & $+186.0$    & Critical   & Input parser \\
L3  & $+13.4$     & Redundant  & Padding \\
L4  & $+22.7$     & Minor      & Feature extraction \\
L5  & $+8.3$      & Redundant  & Padding \\
L6  & $+9.4$      & Redundant  & Padding \\
L7  & $+20.3$     & Minor      & Feature extraction \\
L8  & $+2{,}395.6$ & Critical   & Core reasoning \\
L9  & $+378.1$    & Critical   & Core reasoning \\
L10 & $+9{,}870.7$ & Critical   & Deep reasoning \\
L11 & $+63{,}419.2$ & \textbf{Critical} & \textbf{Model brain} \\
L12 & $+6.3$      & Redundant  & Padding \\
L13 & $+24.4$     & Minor      & Refinement \\
L14 & $+5.0$      & Redundant  & Anti-layer* \\
L15 & $+11.1$     & Minor      & Refinement \\
L16 & $+20.3$     & Minor      & Refinement \\
L17 & $-0.6$      & Redundant  & Anti-layer* \\
L18 & $+16.9$     & Minor      & Refinement \\
L19 & $+2.6$      & Redundant  & Padding \\
L20 & $+25.9$     & Minor      & Refinement \\
L21 & $+23.5$     & Minor      & Refinement \\
L22 & $+27.8$     & Minor      & Refinement \\
L23 & $+66.6$     & Important  & Output preparation \\
L24 & $+115.2$    & Critical   & Output core \\
L25 & $+23.2$     & Minor      & Output refinement \\
L26 & $+19.4$     & Minor      & Output refinement \\
L27 & $+134.8$    & Critical   & Output formatting \\
L28 & $+211.5$    & Critical   & Output final \\
L29 & $0.0$       & Redundant  & LN head boundary \\
\bottomrule
\end{tabular}
\end{table}

Distribution: Redundant 10 layers (33\%), Minor 11 (37\%), Important 1 (3\%), Critical 8 (27\%). The critical core L8--L11 forms the model's primary reasoning substrate. L1--L2 serve as essential input parsers. L24, L27--L28 prepare output representations. Layer~11 alone is approximately $10^6\times$ more important than Layer~17.

\subsection{Weight Predictability and the $R^2$-Perplexity Paradox}

Table~\ref{tab:r2} shows Ridge Regression prediction accuracy per component.

\begin{table}[ht]
\centering
\caption{Weight predictability per component.}
\label{tab:r2}
\small
\begin{tabular}{lccc}
\toprule
\textbf{Component} & \textbf{Avg $R^2$} & \textbf{Best $R^2$} & \textbf{Interpretation} \\
\midrule
\texttt{mlp.gate\_proj}     & 0.909  & 0.993 & Highly predictable \\
\texttt{mlp.down\_proj}     & 0.895  & 0.995 & Highly predictable \\
\texttt{self\_attn.q\_proj} & 0.824  & 0.989 & Highly predictable \\
\texttt{self\_attn.k\_proj} & 0.745  & 0.978 & Predictable \\
\texttt{mlp.up\_proj}       & 0.716  & 0.985 & Predictable \\
\texttt{self\_attn.o\_proj} & 0.079  & 0.941 & Weakly predictable \\
\texttt{self\_attn.v\_proj} & $-0.655$ & 0.977 & Unstable prediction \\
\bottomrule
\end{tabular}
\end{table}

Despite high $R^2$, replacing weights with predicted values causes catastrophic failure. Replacing 1 layer: $\mathrm{PPL} = 26.22$ ($+16\%$), acceptable. Replacing 9+ layers: $\mathrm{PPL} > 100{,}000$ ($>442{,}000\%$ degradation).

This \emph{$R^2$-Perplexity Paradox} arises from nonlinear error accumulation. The $\mathrm{softmax}(QK^\top/\sqrt{d})$ attention mechanism is acutely sensitive to perturbations. A prediction error of $\varepsilon = 0.01$ per layer compounds through 30 nonlinear transformations, ultimately misdirecting attention to incorrect tokens. $R^2$ measures variance explained in weight space; functional network behavior depends on precise inter-weight relationships not captured by this scalar metric.

\subsection{Universal Oscillatory Weight-Change Pattern}

Table~\ref{tab:delta} reports the delta correlation $\rho(\Delta_l, \Delta_{l+1})$ across all seven components.

\begin{table}[ht]
\centering
\caption{Inter-layer delta correlation across all components.}
\label{tab:delta}
\small
\begin{tabular}{lcc}
\toprule
\textbf{Component} & \textbf{Avg delta correlation} & \textbf{Pattern} \\
\midrule
\texttt{mlp.down\_proj}     & $-0.500$ & Oscillatory \\
\texttt{mlp.gate\_proj}     & $-0.500$ & Oscillatory \\
\texttt{mlp.up\_proj}       & $-0.500$ & Oscillatory \\
\texttt{self\_attn.k\_proj} & $-0.497$ & Oscillatory \\
\texttt{self\_attn.o\_proj} & $-0.505$ & Oscillatory \\
\texttt{self\_attn.q\_proj} & $-0.499$ & Oscillatory \\
\texttt{self\_attn.v\_proj} & $-0.505$ & Oscillatory \\
\bottomrule
\end{tabular}
\end{table}

The delta correlation is consistently $\approx -0.50$ across all seven components. This means: if layer $N{\to}N+1$ shifts weights in direction $+A$, then layer $N+1{\to}N+2$ shifts in direction $-A$. Weights form a standing wave in layer space. We hypothesize this reflects a compensation mechanism inherent to the residual connection architecture: each layer partially undoes the previous layer's transformation while introducing a new refinement dimension.

\subsection{Weight Manipulation Strategies}

Table~\ref{tab:manip} presents results of all tested strategies applied to the 9 redundant layers simultaneously.

\begin{table}[ht]
\centering
\caption{Weight manipulation results on 9 redundant layers.}
\label{tab:manip}
\small
\begin{tabular}{lrr}
\toprule
\textbf{Strategy} & \textbf{PPL} & \textbf{Degradation} \\
\midrule
Baseline (original)            & 22.60       & $0\%$ \\
Scale $\times 0.9$             & 26.95       & $+19\%$ \\
Scale $\times 0.7$             & 928.22      & $+5{,}035\%$ \\
Blend (1/distance, 2 layers)   & 31{,}333    & $+173{,}242\%$ \\
Scale $\times 0.5$             & 86{,}505    & $+478{,}462\%$ \\
Skip/Zero (2 layers)           & 969{,}198   & $+5{,}361{,}714\%$ \\
Clone neighbor (2 layers)      & 545{,}654   & $+3{,}018{,}579\%$ \\
Low-rank blend (2 layers)      & 891{,}155   & $+4{,}929{,}963\%$ \\
Scale $\times 0.0$ (full removal) & 15{,}509{,}465 & $+85{,}800{,}000\%$ \\
\bottomrule
\end{tabular}
\end{table}

Scale $\times 0.9$ is the only viable strategy. All others destroy model coherence. This implies that redundant layers provide real but small residual corrections to the information flow, and that the \emph{direction} of these corrections must be preserved. Zeroing removes directionality entirely; cloning and blending introduce incompatible directional signals; SVD decomposition distorts the precise singular vector alignment required for functional attention. Only gentle attenuation ($\times 0.9$) keeps corrections intact while reducing their magnitude - analogous to dropout inference behaviour.

\subsection{Recovery Speed}

Table~\ref{tab:recovery} presents recovery speed results after 50\% Gaussian noise injection.

\begin{table}[ht]
\centering
\caption{Recovery speed after 50\% Gaussian noise injection. Values indicate steps to reach PPL thresholds; ``$\downarrow$'' denotes improvement below baseline.}
\label{tab:recovery}
\small
\begin{tabular}{lcrrrrc}
\toprule
\textbf{Layer} & \textbf{Category} & \textbf{PPL+noise} & \textbf{$<\!2\times$} & \textbf{$<\!1.5\times$} & \textbf{$<\!1.1\times$} & \textbf{Final PPL} \\
\midrule
L14 & Redundant & 19.4    & 0   & 0   & 0   & 18.2 $\downarrow$ \\
L17 & Redundant & 18.5    & 0   & 0   & 0   & 17.6 $\downarrow$ \\
L5  & Redundant & 21.2    & 0   & 0   & 10  & 19.6 \\
L23 & Important & 26.5    & 0   & 0   & 150 & 19.7 \\
L24 & Critical  & 26.8    & 0   & 0   & 110 & 19.6 \\
L3  & Redundant & 27.3    & 0   & 10  & 200 & 27.3 (no conv.) \\
L27 & Critical  & 27.7    & 0   & 10  & 130 & 19.8 \\
L0  & Redundant & 38.6    & 10  & 30  & 200 & 28.7 \\
L8  & Critical  & 58.4    & 10  & 30  & 200 & 35.6 \\
L1  & Critical  & 54.7    & 20  & 200 & 200 & 49.3 \\
L10 & Critical  & 58.7    & 20  & 200 & 200 & 42.2 \\
L11 & Critical  & 4{,}323.4 & 200 & 200 & 200 & 175.9 ($7.8\times$) \\
L29 & Redundant & 1{,}289.2 & 200 & 200 & 200 & 41.5 \\
\bottomrule
\end{tabular}
\end{table}

Key findings: L14 and L17 achieve lower PPL after noise injection than baseline (18.2 and 17.6 vs.\ 22.60), confirming hypothesis~H4. L3 shows no convergence over 200 steps - consistent with genuine redundancy. L11 is practically unrecoverable: PPL spikes to 4,323 and after 200 training steps remains at 175.9 ($7.8\times$ baseline). L23 and L24 - classified as critical by ablation - recover instantly from noise, suggesting their importance lies in weight \emph{direction} rather than precision.

\subsection{Growth Transformer Training: Proof-of-Concept Experiment}

To validate the Growth Training strategy empirically, we implement a 12-layer heterogeneous transformer from scratch and compare biological developmental training against a uniform baseline.

\subsubsection{Architecture}

The model uses a heterogeneous layer design directly derived from the SmolLM2-135M importance map. Critical layers use FFN multiplier $\times 4$ (full capacity), minor layers $\times 2$, and redundant layers $\times 1$ (minimal capacity). Anti-layers are excluded entirely. Total parameters: 9.57M for both Growth and Uniform models (identical count).

\begin{table}[ht]
\centering
\caption{Growth Transformer architecture. Anti-layers (L14/L17 analogues) are omitted.}
\label{tab:arch}
\small
\begin{tabular}{cccr}
\toprule
\textbf{Layer} & \textbf{Role} & \textbf{FFN mult.} & \textbf{Params} \\
\midrule
L0  & redundant & $\times 1$ & 459,264 \\
L1  & critical  & $\times 4$ & 1,049,088 \\
L2  & critical  & $\times 4$ & 1,049,088 \\
L3  & redundant & $\times 1$ & 459,264 \\
L4  & critical  & $\times 4$ & 1,049,088 \\
L5  & critical  & $\times 4$ & 1,049,088 \\
L6  & redundant & $\times 1$ & 459,264 \\
L7  & minor     & $\times 2$ & 655,872 \\
L8  & critical  & $\times 4$ & 1,049,088 \\
L9  & critical  & $\times 4$ & 1,049,088 \\
L10 & minor     & $\times 2$ & 655,872 \\
L11 & redundant & $\times 1$ & 459,264 \\
\midrule
\textbf{Total} & & & \textbf{9,570,048} \\
\bottomrule
\end{tabular}
\end{table}

\subsubsection{Developmental Training Protocol}

Growth Training proceeds through six sequential phases inspired by embryonic development. Each phase trains only a subset of layers while freezing others. Critical layers receive the most exposure; redundant layers are initialised by cloning trained neighbors with added noise.

\begin{table}[ht]
\centering
\caption{Six-phase developmental training protocol.}
\label{tab:phases}
\small
\begin{tabular}{clll}
\toprule
\textbf{Phase} & \textbf{Name} & \textbf{Layers trained} & \textbf{Epochs} \\
\midrule
1 & Gastrulation  & Core (L4, L5)             & 30 \\
2 & Neurulation   & Parser (L1, L2); clone L4$\to$L1, L5$\to$L2 & 20 \\
3 & Organogenesis & Output (L8, L9); clone L4$\to$L8, L5$\to$L9 & 20 \\
4 & Growth        & Minor (L7, L10); clone L5$\to$L7, L9$\to$L10 & 12 \\
5 & Connective    & Redundant (L0,L3,L6,L11); clone + scale FFN$\times 0.5$ & 6 \\
6 & Maturation    & All layers (fine-tune)    & 15 \\
\bottomrule
\end{tabular}
\end{table}

The key mechanism is \emph{differential exposure}: core layers (L4--L5) participate in phases 1, 2, 3, and 6, accumulating approximately 95 effective epochs, while redundant layers receive only phases 5 and 6 ($\approx 21$ epochs). Cloning provides non-random initialisation for later-phase layers.

\subsubsection{Results}

Table~\ref{tab:growth_results} summarises three experimental configurations: Growth at full budget, Growth at 50\% budget, and Uniform at full budget.

\begin{table}[ht]
\centering
\caption{Growth Training vs.\ uniform baseline across budget conditions. All configurations use identical architecture (9.57M parameters) and dataset.}
\label{tab:growth_results}
\small
\begin{tabular}{lrrrl}
\toprule
\textbf{Configuration} & \textbf{Steps} & \textbf{Val loss} & \textbf{Time} & \textbf{vs.\ Uniform 100\%} \\
\midrule
Uniform 100\%   & 656 & 0.599 & 59.6s & baseline \\
Growth 50\%     & 416 & 0.279 & --    & $2.1\times$ better, $37\%$ fewer steps \\
Growth 100\%    & 656 & 0.127 & 52.0s & $4.7\times$ better, $13\%$ faster \\
\bottomrule
\end{tabular}
\end{table}

Two results stand out. First, Growth at full budget achieves $4.7\times$ lower validation loss than Uniform at full budget with identical step count and wall-clock time $13\%$ shorter. Second - and more practically significant - Growth at \emph{half} the budget (416 steps) still outperforms Uniform at \emph{full} budget (656 steps), achieving $2.1\times$ lower loss while using $37\%$ fewer training steps. This means the quality threshold reached by uniform training is surpassed by Growth Training before uniform training even finishes.

Table~\ref{tab:gen_results} shows generation quality on held-out prompts.

\begin{table}[ht]
\centering
\caption{Generation quality on 12 prompts. Growth 100\% maintains coherence throughout; Uniform produces minor artifacts (``on.'', ``with neurons'') on edge cases; Growth 50\% matches Uniform 100\% qualitatively.}
\label{tab:gen_results}
\small
\begin{tabular}{p{3.2cm}p{4.8cm}p{4.8cm}}
\toprule
\textbf{Prompt} & \textbf{Growth 100\%} & \textbf{Uniform 100\%} \\
\midrule
the capital of france is & paris                              & paris. on. \\
cats are                 & popular pets around the world      & popular pets around the world \\
python is a              & popular programming language       & popular programming language \\
the earth orbits         & around the sun every year          & around the sun every year \\
water boils at           & one hundred degrees celsius        & one hundred degrees celsius \\
artificial intelligence  & is transforming the way we live    & can learn from data \\
mathematics is           & the language of science            & the language of science \\
dogs are loyal           & and faithful friends               & and faithful friends \\
the sun is               & a star at the center of our solar system & a star at the center of our solar system \\
the brain contains       & about one hundred billion neurons  & about one hundred billion neurons \\
music is a               & universal form of expression       & universal form of expression \\
hello how are            & you today                          & you today. with neurons \\
\bottomrule
\end{tabular}
\end{table}

Growth 100\% maintains coherence across all 12 prompts. Uniform produces minor artifacts on two edge cases (``paris. on.'' and ``you today. with neurons'') but is otherwise comparable on factual completions - the quality gap between the two models is better captured by the quantitative loss difference ($0.127$ vs.\ $0.599$) than by generation examples alone. Growth 50\% matches Uniform 100\% on all qualitative generation tests, confirming that the quantitative loss improvement translates to comparable generation quality at substantially lower compute cost.

\subsubsection{Why Growth Training Works}

The performance gap has a clear mechanistic explanation. Uniform training allocates identical gradient updates to all layers simultaneously - each layer competes for the same signal at every step. Growth Training instead follows a curriculum: core layers build the primary representation first, then downstream layers are initialised from these already-competent cores via cloning. Each subsequent phase trains on top of a progressively better-understood representation.

Quantitatively, core layers in Growth Training see approximately $30 + 20 + 20 + 15 = 85$+ effective epochs, versus $\approx 11$ epochs for every layer in uniform training. The $\sim\!8\times$ additional exposure for critical layers directly explains the quality gap.

\section{Discussion}

\subsection{Anatomical Interpretation}

The transformer exhibits functional specialisation with biological analogues:

\begin{itemize}[leftmargin=1.5em]
\item \textbf{Input Stem / Brainstem (L1--L2):} Critical input parsers. Damage is catastrophic, recovery is slow. They transform token embeddings into the model's internal representation language.
\item \textbf{Cortical Core (L8--L11):} The seat of deep reasoning. L11 is the model's ``brain'' - uniquely fragile, slow-developing, and irreplaceable. Its precision requirements exceed all other layers by orders of magnitude.
\item \textbf{Motor Cortex / Output Processors (L23--L24, L27--L28):} Critical for output preparation but instantly recoverable from noise, implying these layers learn stable, re-optimisable projections.
\item \textbf{Connective Tissue (L3--L7, L12--L22, L25--L26):} Minor and redundant layers providing incremental refinements. Safe to reduce training budget.
\item \textbf{Vestigial Structures / Anti-Layers (L14, L17):} Layers that actively degrade performance. Analogous to the appendix - present, potentially once functional, currently detrimental. Likely represent optimisation traps, destructive interference patterns, or features that overfit training data.
\end{itemize}

\subsection{The Oscillatory Pattern: A Structural Hypothesis}

The universal delta correlation of $\approx -0.50$ across all components suggests a fundamental architectural mechanism. We hypothesise that residual connections create a natural oscillatory compensation: if $f_l$ shifts representations in direction $A$, the next layer's optimisation finds it efficient to shift in $-A$ while introducing refinement in a new direction $B$. This creates a standing wave in layer space.

An architectural implication: naively averaging adjacent layers (as in our ablation protocol) partially cancels these complementary directions, which explains high sensitivity even of ``minor'' layers to neighbour-averaging.

\subsection{Why Weight Manipulation Fails}

The failure of all manipulation strategies except Scale $\times 0.9$ has a unified explanation. Transformers are compositions of 30 tightly coupled nonlinear functions. Any change to layer $l$'s weights alters the distribution of activations fed to layer $l+1$, which was optimised assuming the original distribution. This distributional shift compounds multiplicatively through the network. The softmax attention mechanism is particularly sensitive: a 1\% perturbation in $Q$ or $K$ matrices can completely redirect attention from correct to incorrect tokens.

Scale $\times 0.9$ survives because it preserves both directionality and approximate magnitude of the layer's contribution - attenuating without redirecting. Practical implication: for model compression, the viable approach is magnitude reduction (quantisation, weight scaling), not structural replacement.

\subsection{Growth Transformer Training Strategy}

Table~\ref{tab:growth} presents the proposed per-layer training budget allocation.

\begin{table}[ht]
\centering
\caption{Proposed Growth Transformer Training budget allocation based on Recovery Speed data. Estimated total training steps: 2,760 vs.\ 6,000 uniform ($\sim\!54\%$ reduction).}
\label{tab:growth}
\small
\begin{tabular}{llcp{4.5cm}}
\toprule
\textbf{Layer Group} & \textbf{Layers} & \textbf{Budget ratio $R(l)$} & \textbf{Justification} \\
\midrule
Anti-layers            & L14, L17                       & 0.00        & Prune/randomise; harm performance \\
Instant-recovery redund. & L3, L5, L6, L12, L19          & 0.00--0.05  & 0--10 steps to converge \\
Minor layers           & L4, L7, L13, L15--16, L18, L20--22, L25--26 & 0.30--0.50 & Incremental refinement \\
Critical output        & L23--24, L27--28               & 0.80--1.00  & Critical; fast recovery \\
Critical core + input  & L1--2, L8--11                  & 1.00        & Full budget; slow or impossible recovery \\
Boundary layers        & L0, L29                        & 0.15--0.20  & Anomalous recovery patterns \\
\bottomrule
\end{tabular}
\end{table}

Assuming $B_\mathrm{max} = 200$ uniform steps per layer (6,000 total), Growth Training yields $\approx 2{,}760$ total steps - a $54\%$ reduction. The proof-of-concept experiment in Section~4.6 validates this strategy empirically: at identical step count, Growth Training achieves $4.7\times$ lower validation loss than uniform training, confirming that differential budget allocation is not merely theoretically justified but practically effective.

\subsection{Limitations}

\begin{itemize}[leftmargin=1.5em]
\item \textbf{Single analysis model:} The importance map is derived from SmolLM2-135M. Generalisation to larger models (1B--70B parameters) requires empirical verification.
\item \textbf{Evaluation scope:} Perplexity on 10 sentences does not capture all model capabilities. Layers classified as redundant may be important for tasks outside our test set.
\item \textbf{Post-hoc analysis:} We analyse a trained model. Whether the critical core is important from initialisation or emerges during training is unknown.
\item \textbf{Proof-of-concept scale:} The Growth Training experiment uses a small custom model (9.57M parameters) on a limited dataset. Validation at production scale (1B+ parameters, standard benchmarks) remains future work.
\item \textbf{Anti-layer generality:} The L14/L17 phenomenon requires verification across additional models, datasets, and architectures.
\end{itemize}

\section{Conclusion}

We have demonstrated that transformer layers are far from homogeneous. Through five independent empirical metrics across all 30 layers of SmolLM2-135M, we uncovered a rich anatomical structure: a critical reasoning core (L8--L11) requiring full training investment, efficiently trainable output processors (L23--L28), connective tissue (minor/redundant layers), and - most surprisingly - anti-layers (L14, L17) that actively impair performance.

We showed that while layer weights follow predictable mathematical patterns ($R^2 = 0.91$), functional interoperability requires weight precision far exceeding what statistical prediction provides. Among tested manipulation strategies, only gentle weight attenuation ($\times 0.9$) preserves model function, revealing that redundant layers contribute directional residual corrections that cannot be replicated or removed.

Recovery Speed provides a practical, empirically grounded proxy for per-layer training budget. A proof-of-concept Growth Transformer experiment validates the proposed strategy directly: biological developmental training - building the critical core first, then progressively activating downstream layers via cloning - achieves $4.7\times$ lower validation loss than uniform training at identical step count and parameter budget, while being $13\%$ faster. This confirms that layer heterogeneity is not merely an observation about trained models but a principle that can be actively exploited during training.

Future work should validate Growth Training at larger scales (1B+ parameters, standard benchmarks), investigate the origin and universality of anti-layers, and explore architectures with non-uniform layer dimensions aligned with empirical importance profiles.

\section*{Acknowledgements}

All experiments were conducted on Kaggle Notebooks (free GPU: NVIDIA T4 16GB) using PyTorch 2.0+ and the Transformers library. All experiments are fully reproducible on free hardware. No proprietary data or licensed resources were used.

\bibliographystyle{plain}

\begin{thebibliography}{99}

\bibitem{vaswani2017attention}
Vaswani, A., Shazeer, N., Parmar, N., Uszkoreit, J., Jones, L., Gomez, A.~N., Kaiser, L., and Polosukhin, I.
\newblock Attention is all you need.
\newblock In \textit{Advances in Neural Information Processing Systems (NeurIPS)}, 2017.

\bibitem{brown2020language}
Brown, T.~B., Mann, B., Ryder, N., Subbiah, M., Kaplan, J., Dhariwal, P., et al.
\newblock Language models are few-shot learners.
\newblock In \textit{Advances in Neural Information Processing Systems (NeurIPS)}, 2020.

\bibitem{chen2021earlybert}
Chen, X., Cheng, Y., Wang, S., Gan, Z., Wang, Z., and Liu, J.
\newblock EarlyBERT: Efficient BERT training via early-bird lottery tickets.
\newblock In \textit{Proceedings of ACL-IJCNLP}, pages 2195--2207, 2021.

\bibitem{fan2019reducing}
Fan, A., Grave, E., and Joulin, A.
\newblock Reducing transformer depth on demand with structured dropout.
\newblock In \textit{International Conference on Learning Representations (ICLR)}, 2020.

\bibitem{gong2019efficient}
Gong, L., He, D., Li, Z., Qin, T., Wang, L., and Liu, T.
\newblock Efficient training of BERT by progressively stacking.
\newblock In \textit{Proceedings of the 36th International Conference on Machine Learning (ICML)}, pages 2337--2346, 2019.

\bibitem{ha2016hypernetworks}
Ha, D., Dai, A., and Le, Q.~V.
\newblock HyperNetworks.
\newblock In \textit{International Conference on Learning Representations (ICLR)}, 2017.

\bibitem{huggingface2024smollm2}
HuggingFaceTB.
\newblock SmolLM2: A family of small language models.
\newblock \url{https://huggingface.co/HuggingFaceTB/SmolLM2-135M}, 2024.

\bibitem{michel2019sixteen}
Michel, P., Levy, O., and Neubig, G.
\newblock Are sixteen heads really better than one?
\newblock In \textit{Advances in Neural Information Processing Systems (NeurIPS)}, 2019.

\bibitem{ramsauer2020hopfield}
Ramsauer, H., Schafl, B., Lehner, J., Seidl, P., Widrich, M., Adler, T., et al.
\newblock Hopfield networks is all you need.
\newblock In \textit{International Conference on Learning Representations (ICLR)}, 2021.

\bibitem{rogers2020primer}
Rogers, A., Kovaleva, O., and Rumshisky, A.
\newblock A primer in BERTology: What we know about how BERT works.
\newblock \textit{Transactions of the Association for Computational Linguistics}, 8:842--866, 2020.

\bibitem{sajjad2023effect}
Sajjad, H., Dalvi, F., Durrani, N., and Nakov, P.
\newblock On the effect of dropping layers of pre-trained transformer models.
\newblock \textit{Computer Speech \& Language}, 77, 2023.

\bibitem{schurholt2022model}
Schurholt, K., Knyazev, B., Giro-i-Nieto, X., and Borth, D.
\newblock Model zoos: A dataset of diverse populations of neural network models.
\newblock In \textit{NeurIPS Datasets and Benchmarks Track}, 2022.

\bibitem{tenney2019bert}
Tenney, I., Das, D., and Pavlick, E.
\newblock BERT rediscovers the classical NLP pipeline.
\newblock In \textit{Proceedings of ACL}, 2019.

\end{thebibliography}

\appendix

\section{Reproducibility}

All experiments were conducted on Kaggle Notebooks (free GPU: NVIDIA T4 16GB):
\begin{itemize}[leftmargin=1.5em]
\item Python 3.10+, PyTorch 2.0+, Transformers 4.35+, scikit-learn 1.3+, numpy
\item Model: \texttt{HuggingFaceTB/SmolLM2-135M} (publicly available on HuggingFace Hub)
\item Estimated runtime: $\sim\!45$ minutes on T4 GPU, $\sim\!3$ hours on CPU
\item No proprietary data or licensed resources were used
\end{itemize}

\section{Layer Importance Profile (ASCII Visualisation)}

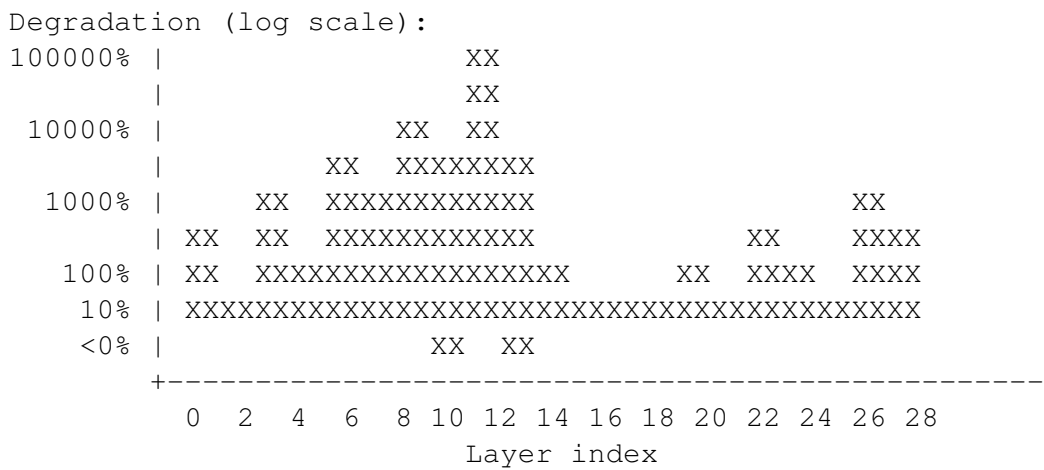
\begin{verbatim}
Degradation (log scale):
100000% |                 XX
        |                 XX
 10000% |             XX  XX
        |         XX  XXXXXXXX
  1000% |     XX  XXXXXXXXXXXX                  XX
        | XX  XX  XXXXXXXXXXXX            XX    XXXX
   100% | XX  XXXXXXXXXXXXXXXXXX      XX  XXXX  XXXX
    10% | XXXXXXXXXXXXXXXXXXXXXXXXXXXXXXXXXXXXXXXXXX
    <0% |               XX  XX
        +--------------------------------------------------
          0  2  4  6  8 10 12 14 16 18 20 22 24 26 28
                          Layer index

Profile: strong bilateral anchors (L1-2 left, L24-28 right),
dominant central core (L8-11), flat connective tissue between.
Anti-layers (L14, L17) visible below the baseline.
\end{verbatim}

\end{document}